\documentclass[runningheads]{llncs}
\usepackage{graphicx}
\usepackage[titles]{tocloft}
\usepackage{amsmath,amssymb,kotex,color}
\usepackage{tikz}
\usetikzlibrary{shapes,arrows}
\usepackage{setspace}
\usepackage{array}
\usepackage{romanbar}
\usepackage{pgfplots}
\usepackage{caption, subcaption}
\usepackage{booktabs}
\usepackage{threeparttable}
\usepackage{makecell, multirow, cellspace}
\usepackage{rotating}
\usepackage{comment}
\usepackage{xurl}
\usepackage{hyperref}

\begin{document}
\title{6MapNet: Representing Soccer Players from Tracking Data by a Triplet Network}
\titlerunning{6MapNet: Representing Soccer Players by a Triplet Network}
%
\author{Hyunsung Kim\inst{1}\orcidID{0000-0002-6286-5160} \and
Jihun Kim\inst{2} \and
Dongwook Chung\inst{1} \and \\
Jonghyun Lee\inst{1} \and
Jinsung Yoon\inst{1} \and
Sang-Ki Ko\inst{1,3}}
\authorrunning{H. Kim et al.}
%
\institute{Fitogether Inc., Seoul, South Korea \and
Seoul National University, Seoul, South Korea \and
Kangwon National University, Chuncheon, South Korea}
\maketitle
\begin{abstract}
Although the values of individual soccer players have become astronomical, subjective judgments still play a big part in the player analysis. Recently, there have been new attempts to quantitatively grasp players' styles using video-based event stream data. However, they have some limitations in scalability due to high annotation costs and sparsity of event stream data. In this paper, we build a triplet network named 6MapNet that can effectively capture the movement styles of players using in-game GPS data. Without any annotation of soccer-specific actions, we use players' locations and velocities to generate two types of heatmaps. Our subnetworks then map these heatmap pairs into feature vectors whose similarity corresponds to the actual similarity of playing styles. The experimental results show that players can be accurately identified with only a small number of matches by our method.
\end{abstract}
\section{Introduction}
The values of individual players have become astronomical with the growth of the soccer industry. Nevertheless, the absence of objective criteria for understanding players causes a considerable gap between player price and utility. As a result, cheaply scouted players often hit the jackpot, while players signed with a huge transfer fee sometimes fail. Thus, scouts have spent a lot of time and effort watching games of unknown players to find hidden gems without wasting money.

Recently, there have been attempts to quantitatively grasp players' styles using video-based event stream data~\cite{Decroos2019,Decroos2020,Gyarmati2016}, but they have some limitations due to the properties of the data. First, since the event data were too sparse to figure out a player's style from a few matches, they used the seasonal aggregations of individual players. This causes difficulty meeting the demand of real-world scouts who do not have an unknown player's full-season data. Moreover, they did not consider that a player could take multiple tactical positions or roles during a season, which significantly affects their locational tendency. Above all, the data collection requires much manual work by human annotators. Representing playing styles prove its worth when applied to youth or lower leagues for scouting, while these annotation costs make these studies hard to be widespread.

This paper proposes a triplet network named 6MapNet to find a robust representation of playing styles from in-game GPS data. As the playing style is a subjective concept, there is no label like ``These players have similar playing styles.'' enabling supervised learning. Instead, we construct a semi-supervised learning framework with the label ``These inputs have similar playing styles because they are the same player's data taking the same tactical role.'' Specifically, we use players' locations and velocities to compute their location and direction heatmaps, respectively. Roles are represented and assigned to players per ``phase'' in matches using the method proposed by Bialkowski et al.~\cite{Bialkowski2014}. We label the pair of a player's phase data as ``similar'' only if the player has taken the same role in both phases. Inspired by FaceNet~\cite{Schroff2015}, we then build a siamese architecture with triplet loss to make the similarity between the feature vectors implies that between the players' actual playing styles.

We evaluate our method by performing a player identification task proposed by Decroos et al.~\cite{Decroos2019}, but with a newly defined likelihood-based similarity named ATL-sim. The experimental results show that our method accurately identifies the anonymized players with only a few matches.

The main contributions of our paper are as follows.
\begin{enumerate}
\item[(a)] Instead of manually annotated event stream data, automatically collected locations and velocity vectors are used to capture playing style.
\item[(b)] Our proposed method is data-efficient in that it can identify a player with only a few matches without using a seasonal aggregation of data.
\item[(c)] We take players' roles into account, which is more reasonable in that a player's movements highly depend on the role given for the match.
\end{enumerate}

\section{Related Works}
\subsection{Playing Style Representation in Soccer} \label{related_works1}
With the acquisition of large-scale sports data, several studies have tried to quantitatively characterize~\cite{Decroos2019,Decroos2020,Gyarmati2016,Pena2015} or evaluate~\cite{Brooks2016,Decroos2019a,Duch2010,Luo2020,Pappalardo2019} soccer players using those data. Especially, some of them used locations (and directions) of actions to represent each player as a vector. Gyarmati et al.~\cite{Gyarmati2016} clustered the on-the-ball actions by their start and end locations and made a feature vector for each player by simply counting the events per cluster. Decroos et al.~\cite{Decroos2019} found the principal components of each action type by applying non-negative matrix factorization (NMF) to the action heatmaps. Players' feature vectors were then assembled by concatenating the weights multiplied on the components to reconstruct the original heatmaps. Decroos et al.~\cite{Decroos2020} also proposed a mixture model that fits the distribution of each action type as a combination of finite Von Mises distributions by the locations and directions of actions. Then, they represented each action as the vector of responsibilities for those component distributions.

\subsection{Siamese Neural Networks and Triplet Loss} \label{related_works2}
The siamese network is a neural network with multiple subnetworks sharing weights. It learns a representation from pairs of images by decreasing the distance between like pairs and increasing that between unlike ones. It was first proposed by Bromley et al.~\cite{Bromley1993} and embodied by Chopra et al.~\cite{Chopra2005}. Koch et al.~\cite{Koch2015} showed the power of the siamese network in one-shot image recognition.

Schroff et al.~\cite{Schroff2015} introduced FaceNet, an extended version of the siamese network. They composed triplets from the dataset with two images of the same label and one of another label, respectively. The triplet loss was then minimized by training the triplet network so that the distance between the former pair (named ``positive pair'') became closer than another (named ``negative pair'').

In this study, we employ the structure of FaceNet to represent playing styles since it has less strict loss than the original siamese network. That is, the triplet loss remains small as long as the positive pair is closer than the negative pair. As different players (i.e., with distinct labels) can actually have similar playing styles, we think the triplet loss is suitable for our problem.

\section{Learning Approach}
In this section, we illustrate the process of playing style representation. First, we describe the data preparation step in Section~\ref{data_preparation} and the labeling step using the role representation in Section~\ref{data_labeling}. The location and direction heatmaps are generated in Section~\ref{heatmap_gen} and then augmented in Section~\ref{heatmap_aug} by simple pixel-wise additions. In Section~\ref{6mapnet}, we introduce a triplet network named 6MapNet that makes the embeddings closer if they are of the same player taking the same role.

\subsection{Data Preparation} \label{data_preparation}
During soccer matches, wearable GPS devices (OhCoach Cell B developed by Fitogether) are attached to the players' upper backs and track their movements. The devices have successfully recorded the latitudes, longitudes, and speeds by 10 Hz from 750 matches of 2019 and 2020 K League 1 and 2, two seasons of the South Korean professional soccer league divisions. (Note that both teams used the devices for some matches, and only one team did for other matches. As such, to avoid confusion, we count twice a match in which both teams are measured.)

After that, we transform the raw locations into the x and y coordinates relative to the pitch. The data from one of the two halves are rotated to maintain the attacking direction. Also, we calculate the 2-dimensional velocities by differentiating the x and y coordinates.

Meanwhile, we divide each match into several {\em phases} in each of which the players' roles are assumed to be consistent. Considering that most formation changes occur when a new session starts or player composition changes, we split each match by the half time, player substitution, and dismissal times. (Phases with a length of 10 minutes or less are absorbed into adjacent phases to ensure the minimum duration.) This breaks the whole dataset down into each player's phase-by-phase data, which we call {\em player-phase entities}.

As a result, 635 matches are divided into 1,989 phases having total of 17,953 player-phase entities generated from 436 players. Each player-phase entity has a time series of locations $(s_x, s_y)$ in meters and velocities $(v_x, v_y)$ in m/s.

\subsection{Automated Data Labeling Based on Role Representation} \label{data_labeling}
In this section, labels for the triplet network are automatically generated from phase-by-phase role learning and clustering. For the former, we use the role representation proposed by Bialkowski et al.~\cite{Bialkowski2014}. We define the player-phase entity's role as the most frequent role among the roles assigned frame-by-frame. (Please refer to \cite{Bialkowski2014} for more details. Note that Bialkowski et al. ran their algorithm for every {\em half} of a match, while we do for every {\em phase}.)

\begin{figure}[!htb]
\centering
\subfloat[Donggook Lee \label{fig_clustering1}]{
	\includegraphics[width=.32\textwidth]{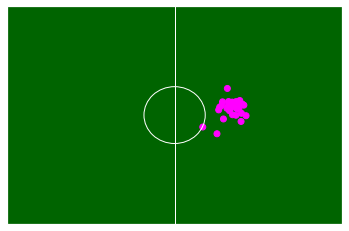}
}
\subfloat[Osmar Ib\'a\~nez \label{fig_clustering2}]{
	\includegraphics[width=.32\textwidth]{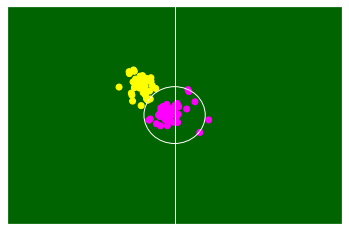}
}
\subfloat[Insung Kim \label{fig_clustering3}]{
	\includegraphics[width=.32\textwidth]{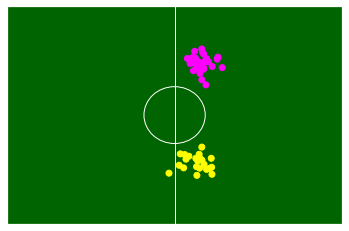}
}\vspace{0.5em}
\subfloat[Yong Lee \label{fig_clustering4}]{
	\includegraphics[width=.32\textwidth]{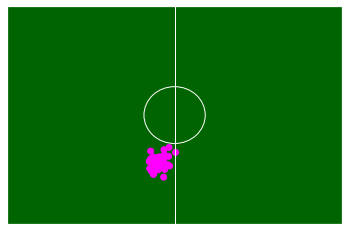}
}
\subfloat[Hyunsoo Hwang \label{fig_clustering5}]{
	\includegraphics[width=.32\textwidth]{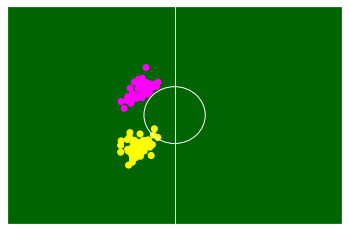}
}
\subfloat[Wanderson Carvalho \label{fig_clustering6}]{
	\includegraphics[width=.32\textwidth]{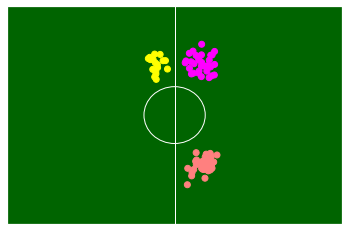}
}
\caption{Mean locations of individual players' roles colored per cluster. Even the same player's data get distinct labels if their colors differ from each other.}
\label{fig_clusterings}
\end{figure}

The outcome of \cite{Bialkowski2014} is a 2D distribution representing a role per player-phase entity. (Note that distributions of the same role vary slightly by phase.) Since our purpose here is to classify the phase data of each player according to the roles, we perform the player-wise clustering on the mean locations of assigned roles to determine which roles to consider as the same or different. To be specific, we apply K-means clustering per player with between two to four clusters maximizing the silhouette score based on the domain knowledge that a player hardly plays more than four positions. If all the scores are less than 0.6, a single cluster is assigned to the player's entire data. See Fig.~\ref{fig_clusterings} as an example.

We then label player-phase entities as pairs of player IDs and role clusters. Namely, if the roles of the same player’s two phases belong to a single cluster, they are labeled as the same identity. On the other hand, if those of the same player belong to different clusters, they are labeled as different identities. We call each of these role-based identities a {\em player-role entity}.

\subsection{Generating Location and Direction Heatmaps} \label{heatmap_gen}
For a labeled time series, two heatmaps are generated by making grids of size $35 \times 50$ on the fixed domains and counting data points in each grid cell. Like in Decroos et al.~\cite{Decroos2019}, we compute each player-phase entity’s location heatmap by making a grid on the soccer pitch. To take into account the running direction, which is not covered by location heatmaps but is still a significant feature, we also compute heatmaps for velocity vectors whose speed is higher than a threshold speed. Like location heatmaps, we overlay a grid on the rectangle
\[
\{ (v_x, v_y) | -12 \text{m/s} \leq v_x \leq 12 \text{m/s}, -8 \text{m/s} \leq v_y \leq 8 \text{m/s} \},
\]
and count the endpoints of the velocity vectors in each grid cell.

\begin{figure}[!htb]
\centering
\subfloat[A location heatmap by making a grid on the soccer pitch and counting the locations in each cell.\label{fig_loc_heatmap}]{
	\includegraphics[width=.4\textwidth]{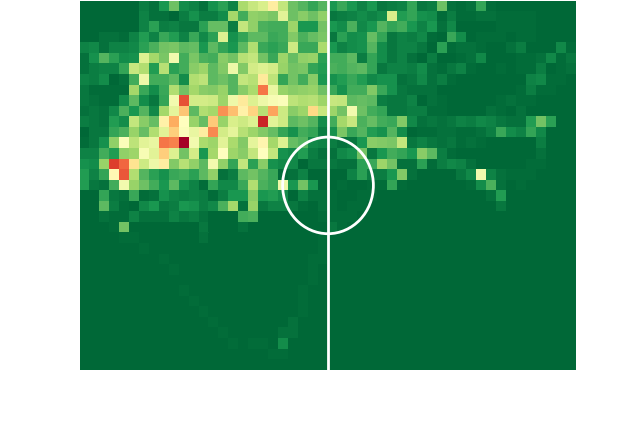}
}\hfil
\subfloat[A direction heatmap by making a grid and counting the endpoints of velocity vectors in each cell.\label{fig_dir_heatmap}]{
	\includegraphics[width=.4\textwidth]{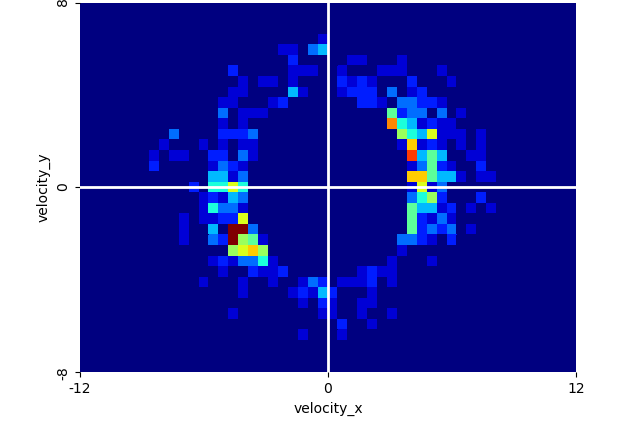}
}
\caption{The location and direction heatmaps of a player-phase entity.}
\label{fig_heatmaps}
\end{figure}

The boundary values of the x-axis and y-axis of the direction heatmap (i.e., $\pm 12 \text{m/s}$ and $\pm 8 \text{m/s}$, respectively) are determined so that the shape of each grid cell is close to a square.
We empirically set the threshold speed as 4m/s. For lower speed thresholds, data points are concentrated near the origin without any tendency. Also, for higher speed thresholds, the number of data points is not enough for robust analysis. Fig.~\ref{fig_heatmaps} shows an example of a pair of heatmaps.

Before the data augmentation, we split the dataset into training, validation, and test sets. For the fair evaluation in Section~\ref{experiments}, the test dataset is constructed by sampling ten heatmap pairs (i.e., pair of location and direction heatmaps) for each of the 308 player-role entities with more than 20 phases in the dataset. Similarly, the validation set is constructed by sampling five phases for each of the 332 entities having more than 15 phases in the remaining data.

\subsection{Data Augmentation by Accumulating Heatmaps} \label{heatmap_aug}
Before to put heatmaps in 6MapNet, we perform the augmentation to make individual heatmaps have enough durations and to enrich the input data. One of the good properties of players’ heatmaps is their ``additivity''. Namely, we can compute the accumulated heatmaps of the multiple time intervals by simple pixel-wise additions. Formally speaking, let $s_p: \mathcal{T} \rightarrow \mathbb{R}^2$ and $v_p: \mathcal{T} \rightarrow \mathbb{R}^2$ be the location and velocity of a player $p$, respectively defined on the time domain $\mathcal{T}$. Also, let $\mathbf{h}: \mathcal{P} \left( \mathbb{R}^2 \right) \rightarrow \mathbb{Z}_{*}^{35 \times 50}$ be the heatmap calculated for a set of 2D vectors. Then for $T = \bigcup_{i=1}^m T_i$ with disjoint time-intervals $T_1, \ldots ,T_m \subset \mathcal{T}$,
\begin{equation}
\mathbf{h}(s_p(T)) = \displaystyle{\sum_{i=1}^m \mathbf{h} (s_p(T_i))}, \quad
\mathbf{h}(v_p(T)) = \displaystyle{\sum_{i=1}^m \mathbf{h} (v_p(T_i))}
\end{equation}
Using this additivity, we augment the dataset by accumulating several heatmaps of the same player-role entities. In other words, we ``crop'' the entire time domain $\mathcal{T}$ of each player-role entity and generate heatmap pairs for the crop.

\begin{figure}[!b]
\centering
\includegraphics[width=\textwidth]{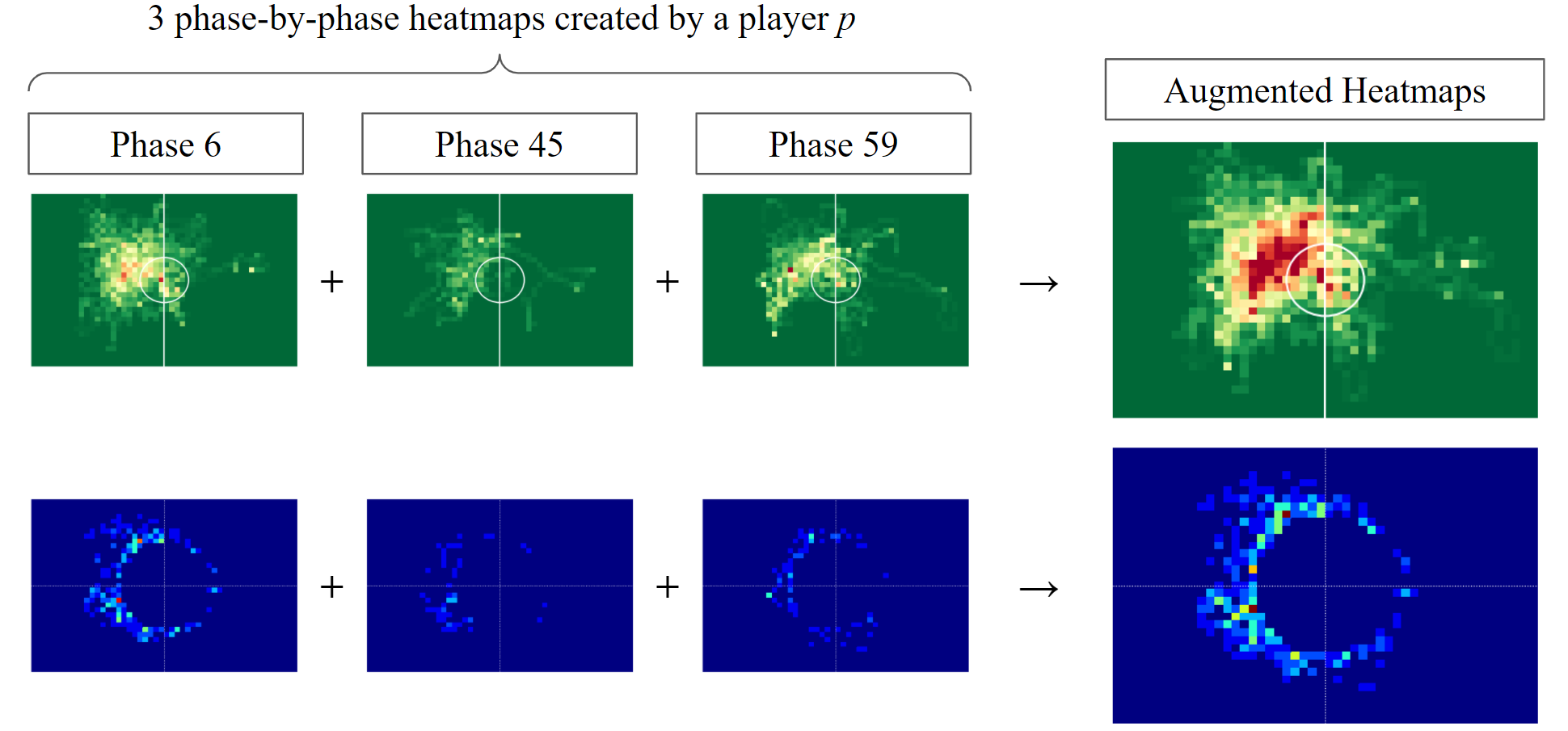}
\caption{An example of heatmap augmentation.}
\label{fig_heatmap_aug}
\end{figure}

The augmentation is performed separately for the training, validation, and test datasets so that data in different sets are not mixed. Three heatmaps are sampled from the set of each player-role entity's heatmaps and added to be a single augmented heatmap. We repeat sampling without duplicates to make a large number of augmented heatmaps. Fig.~\ref{fig_heatmap_aug} illustrates an heatmap augmentation.

Here we choose three as the number of heatmaps from the reasoning below. Too short durations of individual augmented heatmaps are not enough to understand a player's movement tendency. On the other hand, too many heatmaps for one augmentation lead to a high entry barrier for practical use. (i.e., one has to play more matches to have an augmented heatmap.) What we take note of is that three phases of data are collected at once in most cases, since the average duration of a phase is less than 30 minutes. That is, one can usually make a player's 3-combination augmented heatmap by measuring a single full-time match. Hence, we deem the 3-combination as optimal since it both ensures enough durations for analysis and hardly requires more matches comparing to the option of no augmentation.

The way of sampling differs between training and validation/test data. For each of the validation and test datasets, three heatmaps are merged through the exhaustive combination. As such, $\binom{5}{3} = 10$ and $\binom{10}{3} = 120$ accumulated heatmap pairs are generated per player-role entity for the validation and test dataset, respectively. On the other hand, since the number of heatmap pairs for each player-role entity varies in the training dataset, the exhaustive combination deepens the class imbalance. To prevent this, we randomly sample 3-combinations by $4 \cdot n_p$ times for each player-role entity $p$ with $n_p$ heatmap pairs and remove the duplicates.

\begin{figure}[!b]
\centering
\includegraphics[width=\textwidth]{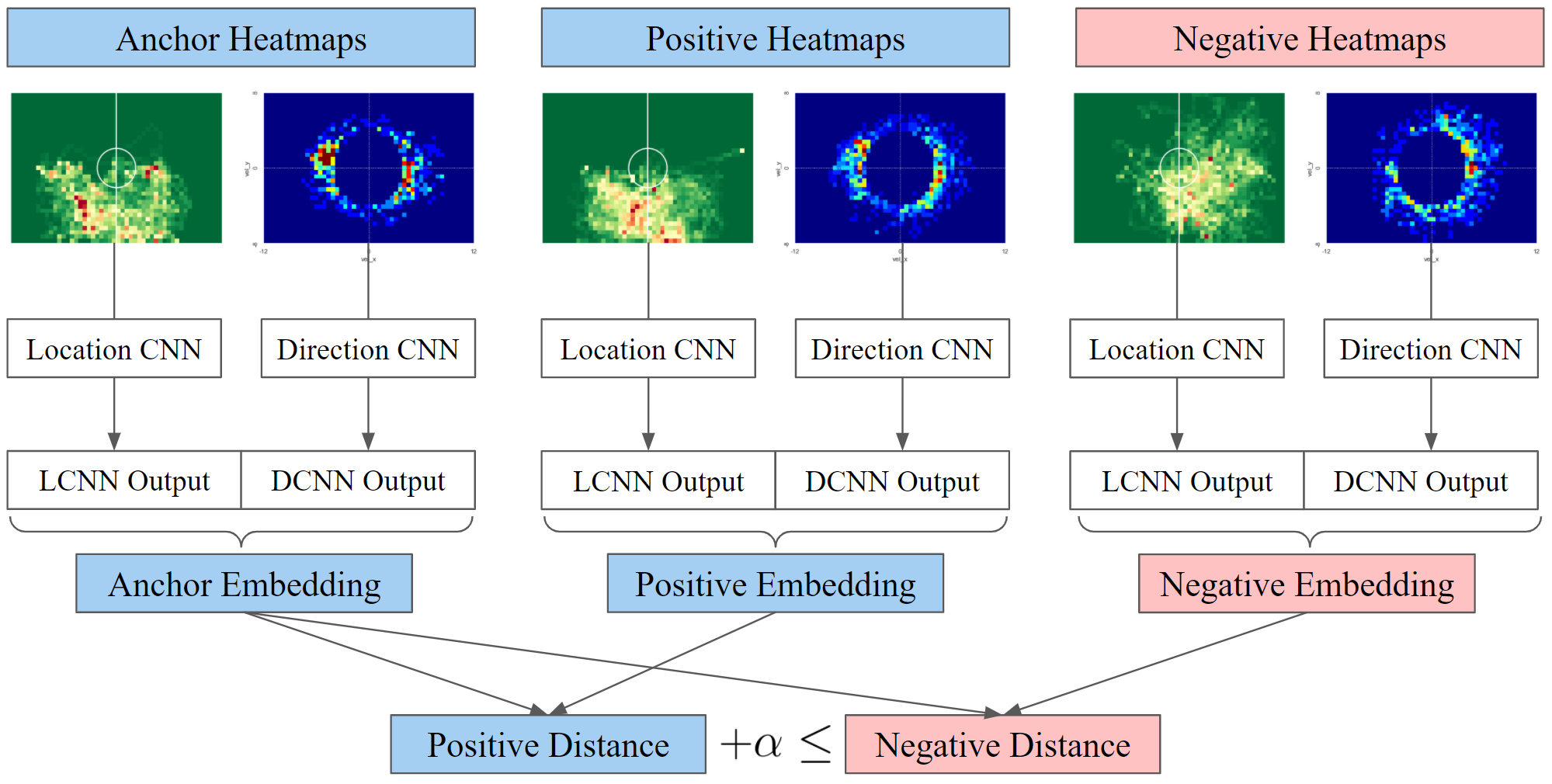}
\caption{6MapNet overview: The anchor and positive heatmap pairs are of the same player-role entity, while the negative is of another entity. The best performance is obtained when $\alpha$ is set to 0.1.}
\label{fig_6mapnet}
\end{figure}

\subsection{Building the Sixfold Heatmap Network \label{6mapnet}}
In this section, we employ the idea of FaceNet~\cite{Schroff2015} and build a triplet network called {\em sixfold heatmap network} (6MapNet) which embeds heatmap pairs of similar players to feature vectors close to each other. FaceNet takes a triplet composed of $\mathbf{x}^a$ (anchor), $\mathbf{x}^p$ (positive), and $\mathbf{x}^n$ (negative) as an input, where $\mathbf{x}^a$ and $\mathbf{x}^p$ are images of the same identity and $\mathbf{x}^n$ is that of another. It learns to find an embedding $f$ that satisfies
\begin{equation}
\Arrowvert f(\mathbf{x}_i^a) - f(\mathbf{x}_i^p) \Arrowvert_2^2 + \alpha \leq \Arrowvert f(\mathbf{x}_i^a) - f(\mathbf{x}_i^n) \Arrowvert_2^2 \label{triplet_ineq}
\end{equation}
by minimizing a triplet loss defined as
\begin{equation} 
\mathcal{L} = \sum_i \left[ \Arrowvert f(\mathbf{x}_i^a) - f(\mathbf{x}_i^p) \Arrowvert_2^2 - \Arrowvert f(\mathbf{x}_i^a) - f(\mathbf{x}_i^n) \Arrowvert_2^2 + \alpha \right]_+ \label{triplet_loss}
\end{equation}

\begin{table}[!t]
\caption{The locational and directional branch CNNs of 6MapNet have the same structure as in this table. It includes the batch normalization after Conv and FC layers and the 25\% dropout after MaxPool layers and Conv4b.} \label{tab_cnn}
\fontsize{9}{10}\selectfont
\centering
\vspace{1em}
{
\renewcommand{\tabcolsep}{0.1cm}
\begin{tabular}{cccccc}
\toprule
Layer & Input & Padding & Kernel & Activation & Output\\
\midrule
Conv1a & $35 \times 50 \times 1$ & $1 \times 0$ & $2 \times 3 \times 4$, 1 & ReLU & $36 \times 48 \times 4$\\
Conv1b & $36 \times 48 \times 4$ & $1 \times 1$ & $3 \times 3 \times 4$, 1 & ReLU & $36 \times 48 \times 4$\\
MaxPool1 & $36 \times 48 \times 4$ & - & $2 \times 2$, 2 & - & $18 \times 24 \times 4$\\
Conv2a & $18 \times 24 \times 4$ & $1 \times 1$ & $3 \times 3 \times 16$, 1 & ReLU & $18 \times 24 \times 16$\\
Conv2b & $18 \times 24 \times 16$ & $1 \times 1$ & $3 \times 3 \times 16$, 1 & ReLU & $18 \times 24 \times 16$\\
MaxPool2 & $18 \times 24 \times 16$ & - & $2 \times 2$, 2 & - & $9 \times 12 \times 16$\\
Conv3a & $9 \times 12 \times 16$ & $1 \times 1$ & $2 \times 3 \times 32$, 1 & ReLU & $10 \times 12 \times 32$\\
Conv3b & $10 \times 12 \times 32$ & $1 \times 1$ & $3 \times 3 \times 32$, 1 & ReLU & $10 \times 12 \times 32$\\
MaxPool3 & $10 \times 12 \times 32$ & - & $2 \times 2$, 2 & - & $5 \times 6 \times 32$\\
Conv4a & $5 \times 6 \times 32$ & $1 \times 1$ & $3 \times 3 \times 64$, 1 & ReLU & $5 \times 6 \times 64$\\
Conv4b & $5 \times 6 \times 64$ & $1 \times 1$ & $3 \times 3 \times 64$, 1 & ReLU & $5 \times 6 \times 64$\\
FC1 & 1920 & - & - & ReLU & 128\\
FC2 & 128 & - & - & - & 10\\
\bottomrule
\end{tabular}}
\end{table}

We apply this architecture by building three identical subnetworks (sharing weights) named 2MapNet, each of which has two branch convolutional neural networks (CNNs). Each subnetwork takes a heatmap pair $\mathbf{x}_i = (\mathbf{h}(s_i), \mathbf{h}(v_i))$ as an input, and each branch CNN (one for the location heatmap $\mathbf{h}(s_i)$ and the other for the direction heatmap $\mathbf{h}(v_i)$) embeds the corresponding heatmap into $\mathbb{R}^d$. After that, we concatenate the two $d$-dimensional outputs and normalize the result so that every embedding $f(\mathbf{x}_i) = f(\mathbf{h}(s_i), \mathbf{h}(v_i)) \in \mathbb{R}^{2d}$ has a unit norm. Consequently, the whole architecture of 6MapNet receives six heatmaps
\[
(\mathbf{x}_i^a, \mathbf{x}_i^p, \mathbf{x}_i^n) = (\mathbf{h}(s_i^a), \mathbf{h}(v_i^a), \mathbf{h}(s_i^p), \mathbf{h}(v_i^p), \mathbf{h}(s_i^n), \mathbf{h}(v_i^n)),
\]
and return a triplet of $2d$-dimensional feature vectors $(f(\mathbf{x}_i^a), f(\mathbf{x}_i^p), f(\mathbf{x}_i^n))$, taking Eq.~\ref{triplet_loss} as a loss to minimize. (See Fig.~\ref{fig_6mapnet} for the overview of the whole architecture, and Table \ref{tab_cnn} for the structure of a single branch CNN.)

As in FaceNet, the choice of triplets has a great influence on the performance. Thus, we decided to select a random positive and a hard negative for each anchor. Specifically, we select triplets as follows for every ten or fewer epochs:
\begin{enumerate}
\item[(a)] Sample five heatmap pairs for each identity to form a candidate set $S$.
\item[(b)] Use every heatmap pair in the dataset as an anchor $\textbf{x}^a$ and combine it with the corresponding five positive candidates $\textbf{x}^p \in S^a = \{ \textbf{x}^p \in S | y^p = y^a \}$ (i.e., $\textbf{x}^p$ with the same label as $\textbf{x}^a$) to make five positive pairs $(\textbf{x}^a, \textbf{x}^p)$.
\item[(c)] For each positive pair $(\textbf{x}^a, \textbf{x}^p)$, pick one hard negative $\textbf{x}^n \in S-S^a$ that does not satisfy Eq.~\ref{triplet_ineq}. If there is no such negative, randomly choose one of the ten negatives with the smallest distance from the anchor.
\end{enumerate}

In particular, the rate of positive pairs with no hard negative in (c), i.e.,
\[
\left| \lbrace (\textbf{x}^a, \textbf{x}^p) \in P : \Arrowvert f(\mathbf{x}_i^a) - f(\mathbf{x}_i^p) \Arrowvert_2^2 + \alpha \leq \Arrowvert f(\mathbf{x}_i^a) - f(\mathbf{x}_i^n) \Arrowvert_2^2 \ \forall \textbf{x}^n \in S-S^a \rbrace \right| / |P|
\]
(where $P$ is the set of all positive pairs selected in (b)) is used as validation accuracy during the training. If there is no improvement in the validation accuracy, our model finishes the whole learning procedure.

We expect that by making the embedding of the same identity (player-role entity) close together, embeddings of similar identities also become close. Because the first triplet selection at the beginning is entirely random, hard negatives can be either similar to their corresponding anchors in terms of playing style or dissimilar to them. For similar anchor-negative pairs, it is difficult to reduce the triplet loss because their similar input heatmaps remain similar after passing through the model. Thus, the model reduces the loss by alienating the dissimilar negatives from their anchors instead. In this way, similar identities remain relatively close, while dissimilar ones move away from each other.

The above inference is verified by the observation that in most cases, the model achieves the best performance just after training the first selected triplets. This implies that the second triplet selection gives a high proportion of hard negatives with similar playing styles to their anchors, making the loss reduction and performance improvement very difficult. Therefore, the resulting model trained only by several epochs with the first triplet selection is expected to return embeddings whose similarities reflect those between actual playing styles.

\section{Experiments} \label{experiments}
Since no objective answer exists for the similarity between players, there is no de-facto standard for evaluating the performance of our method. Thus, we refer to the player identification task proposed by Decroos et al.~\cite{Decroos2019} that checks how well the model identifies the anonymized players using the data from the other season. Based on the assumption that high similarities between the same identities imply those between identities having similar playing styles, we conclude that the high performance of the player identification task implies the success of our method.

To explain, we anonymize the training data defined in Section~\ref{heatmap_gen} and use the test data to de-anonymize them. Having exactly ten phases in the test dataset, each of 308 players has $\binom{10}{3} = 120$ feature vectors as the result of data augmentation in Section~\ref{heatmap_aug}. To take advantage of this plurality, we define and use a likelihood-based similarity instead of $L^p$ norms. We estimate a distribution of training feature vectors per anonymized player-role entity and calculate the log-likelihood that each test feature vector is generated from the distribution. Except for potential outliers, we use the average of $m$ highest values as the similarity among 120 log-likelihoods per player-role entity.

Formally speaking, let $\mathcal{A}$ be the set of player-role entities belonging to both of the training and the test dataset, and $X_{\alpha}^{tr} = \{ \mathbf{x}_i^{tr} \in X^{tr}: y_i^{tr} = \alpha \}$ be the set of the training heatmap pairs for $\alpha \in \mathcal{A}$. We estimate a Gaussian probability density $p_{\alpha}: \mathbb{R}^{2d} \rightarrow \mathbb{R}_+$ using $X_{\alpha}^{tr}$ and calculate the log-likelihoods
\[
   l(\alpha; \mathbf{x}^{te}) = \text{log} \, p_{\alpha}(f(\mathbf{x}^{te}))
\]
of test heatmap pairs $\mathbf{x}^{te} \in X_{\beta}^{te}$ of a labeled player-role entity $\beta \in \mathcal{A}$. Then, for the top-$m$ log-likelihoods $ l(\alpha; \mathbf{x}_{\beta(1)}^{te}), \ldots , l(\alpha; \mathbf{x}_{\beta(M)}^{te})$, we define the {\em average top-$m$ log-likelihood similarity} (ATL-sim) of $\beta$ to $\alpha$ by
\begin{equation}
   \text{sim}(\alpha, \beta; m) = \frac{1}{m} \sum_{i=1}^m l(\alpha; \mathbf{x}_{\beta(i)}^{te}) = \frac{1}{m} \sum_{i=1}^m \text{log} \, p_{\alpha}(f(\mathbf{x}_{\beta(i)}^{te})). 
\end{equation}
Based on this ATL-sim, we evaluated our method by computing the top-$k$ accuracies and mean reciprocal rank (MRR) for varying $m$. That is, for each anonymized entity $\alpha^{tr} \in \mathcal{A}$, we check whether our method correctly found $k$ candidates $\beta_1^{te}, \ldots, \beta_k^{te}$ among the test data such that $\alpha^{tr} \in \{ \beta_1^{te}, \ldots, \beta_k^{te} \}$.

As an ablation study, we compared the results based on different similarity measures and amounts of identifying (test) data. The following is the description of the conditions, where the label `p$_n$-$sim$' indicates the result based on $sim$ using $n$ phases per entity as the identifying data.

\begin{itemize}
	\item \textbf{p$_{10}$-$L^1$}: $L^1$ distance between feature vectors from accumulated heatmaps.
	\item \textbf{p$_{10}$-$L^2$}: $L^2$ distance between feature vectors from accumulated heatmaps.
	\item \textbf{p$_{10}$-AL (p$_{10}$-ATL$_{100}$)}: average log-likelihood without filtering top-$m$.
	\item \textbf{p$_{10}$-ATL$_{75}$}: average top-$75\%$ log-likelihood (top-90 among $\binom{10}{3} = 120$).
	\item \textbf{p$_{10}$-ATL$_{50}$}: average top-$50\%$ log-likelihood (top-60 among $\binom{10}{3} = 120$).
	\item \textbf{p$_6$-ATL$_{25}$}: average top-$25\%$ log-likelihood (top-5 among $\binom{6}{3} = 20$).
	\item \textbf{p$_8$-ATL$_{25}$}: average top-$25\%$ log-likelihood (top-14 among $\binom{8}{3} = 56$).
	\item \textbf{p$_{10}$-ATL$_{25}$}: average top-$25\%$ log-likelihood (top-30 among $\binom{10}{3} = 120$).
	\item \textbf{p$_{10}$-ML}: maximum log-likelihood (top-1 among $\binom{10}{3} = 120$).
\end{itemize}

Table~\ref{tab_performs} shows the results for the player identification task. The best performance is obtained when average top-$25\%$ log-likelihood is used, showing that 46.1\% of the players are correctly identified among 308 players using only ten phases of data (about 289 minutes in total). For most cases (92.5\%), our model correctly suggests ten candidates for identifying the anonymized player.

\begin{table}[!t]
\caption{Top-$k$ accuracies and MRRs of the player identification tasks.} \label{tab_performs}
\fontsize{9}{10}\selectfont
\centering
\vspace{1em}
{\renewcommand{\tabcolsep}{0.1cm}
\begin{tabular}{cccccc}
\toprule
Condition & Top-1 & Top-3 & Top-5 & Top-10 & MRR\\
\midrule
p$_{10}$-$L^1$ & 24.4\% & 44.5\% & 58.4\% & 79.2\% & 0.402\\
p$_{10}$-$L^2$ & 24.7\% & 45.1\% & 59.7\% & 80.2\% & 0.404\\
p$_{10}$-AL & 35.1\% & 59.7\% & 71.4\% & 84.1\% & 0.509\\
p$_{10}$-ATL$_{75}$ & 42.5\% & 66.2\% & 77.6\% & 89.3\% & 0.574\\
p$_{10}$-ATL$_{50}$ & 45.5\% & \textbf{70.5\%} & 80.8\% & 90.6\% & 0.602\\
p$_6$-ATL$_{25}$ & 34.1\% & 60.1\% & 73.7\% & 84.7\% & 0.508\\
p$_8$-ATL$_{25}$ & 41.2\% & 67.2\% & 77.9\% & 89.9\% & 0.574\\
p$_{10}$-ATL$_{25}$ & \textbf{46.1\%} & 69.8\% & \textbf{81.8\%} & \textbf{92.5\%} & \textbf{0.613}\\
p$_{10}$-ML & 37.0\% & 65.9\% & 77.3\% & 91.2\% & 0.547\\
\bottomrule
\end{tabular}}
\end{table}

We also obtained several meaningful observations from the table: (a) p$_{10}$-ATLs outperform p$_{10}$-$L^p$s, meaning that log-likelihoods are more suitable similarity measure for models that generate a plurality of feature vectors per player. (b) The accuracy of p$_n$-ATL improves as $n$ (the number of phases per player-role entity) increases, showing the potential of even better performance as more data is collected. (c) ATL-sim shows the highest performance when $m$ is near 25\% of the total number of phases per player-role entity. The superiority of p$_
n$-ATL$_{25}$ than p$_n$-AL indicates that removing potential outliers enhances the intra-identity consistency of the remaining feature vectors.

We believe our approach to similar player retrieval is more practical in the scouting industry. For example, scouts often want to find cheaper players to replace the core players leaving the teams. Then, player representation models can extract a target list by retrieving similar players to a benchmark player. However, previous approaches need manual annotations to generate event stream data from a whole season to find robust feature vectors representing playing styles. On the other hand, our method does not need any manual work except for putting a device on each player for a few matches to return reliable outputs. This automaticity and data-efficiency can make our method widely used in real-world soccer, including youth or lower divisions.

\section{Conclusion and Future Works}
To our knowledge, this is the first study to characterize soccer players' playing styles using tracking data instead of event stream data. Without using any ball-related information, we use each player's locations and velocities to generate two types of heatmaps. We then build a triplet network named 6MapNet that embeds these heatmap pairs to latent vectors whose similarity reflects the actual similarity of playing styles. The experimental results show that 6MapNet with our newly defined similarity measure ATL-sim can accurately identify anonymized players using data from only a few matches.

In the future, we expect that the use of ``context'' such as sequential information or the movements of teammates and opponents would contribute to the quality of our work. Also, it would justify the choice of model structure to compare ours with some baseline approaches such as principal component analysis or autoencoders. In addition, some practical use cases with real-world examples should be proposed to show the significance of our study. Most of all, we aim to improve the explainability of our model, enabling our work to be understood and practically used by domain participants.

%
%
%
\bibliographystyle{splncs04}
\bibliography{mybibliography}

\begin{thebibliography}{10}
\providecommand{\url}[1]{\texttt{#1}}
\providecommand{\urlprefix}{URL }
\providecommand{\doi}[1]{https://doi.org/#1}

\bibitem{Bialkowski2014}
Bialkowski, A., Lucey, P., Carr, P., Yue, Y., Sridharan, S., Matthews, I.:
  {Large-scale analysis of soccer matches using spatiotemporal tracking data}.
  In: IEEE International Conference on Data Mining (2014)

\bibitem{Bromley1993}
Bromley, J., Guyon, I., LeCun, Y., Sachkinger, E., Shah, R.: {Siamese
  verification using a siamese time delay neural network}. International
  Journal of Pattern Recognition and Artificial Intelligence pp. 737--744
  (1993)

\bibitem{Brooks2016}
Brooks, J., Kerr, M., Guttag, J.: {Developing a data-driven player ranking in
  soccer using predictive model weights}. In: ACM SIGKDD International
  Conference on Knowledge Discovery and Data Mining. vol. 13-17-Augu, pp.
  49--55 (2016)

\bibitem{Chopra2005}
Chopra, S., Hadsell, R., LeCun, Y.: {Learning a similarity metric
  discriminatively, with application to face verification}. In: IEEE Conference
  on Computer Vision and Pattern Recognition (2005)

\bibitem{Decroos2019}
Decroos, T., Davis, J.: {Player vectors: Characterizing soccer players' playing
  style from match event streams}. In: European Conference on Machine Learning
  and Principles and Practice of Knowledge Discovery in Databases (2019)

\bibitem{Decroos2020}
Decroos, T., Roy, M.V., Davis, J.: {SoccerMix: Representing soccer actions with
  mixture models}. In: European Conference on Machine Learning and Principles
  and Practice of Knowledge Discovery in Databases (2020)

\bibitem{Decroos2019a}
Decroos, T., {Van Haaren}, J., Bransen, L., Davis, J.: {Actions speak louder
  than goals: Valuing player actions in soccer}. In: ACM SIGKDD Conference on
  Knowledge Discovery and Data Mining (2019)

\bibitem{Duch2010}
Duch, J., Waitzman, J.S., {Nunes Amaral}, L.A.: {Quantifying the performance of
  individual players in a team activity}. PLoS ONE  \textbf{5}(6) (2010)

\bibitem{Gyarmati2016}
Gyarmati, L., Hefeeda, M.: {Analyzing in-game movements of soccer players at
  scale}. In: MIT Sloan Sports Analytics Conference (2016)

\bibitem{Koch2015}
Koch, G., Zemel, R., Salakhutdinov, R.: {Siamese neural networks for one-shot
  image recognition}. In: International Conference on Machine Learning (2015)

\bibitem{Luo2020}
Luo, Y., Schulte, O., Poupart, P.: {Inverse reinforcement learning for team
  sports: Valuing actions and players}. In: International Joint Conference on
  Artificial Intelligence (2020)

\bibitem{Pappalardo2019}
Pappalardo, L.: {PlayeRank: Data-driven performance evaluation and player
  ranking in soccer via a machine learning approach}. ACM Transactions on
  Intelligent Systems and Technology  \textbf{10}(5) (2019)

\bibitem{Pena2015}
Pe{\~{n}}a, J.L., Navarro, R.S.: {Who can replace Xavi? A passing motif
  analysis of football players}  (2015), \url{http://arxiv.org/abs/1506.07768}

\bibitem{Schroff2015}
Schroff, F., Kalenichenko, D., Philbin, J.: {FaceNet: A unified embedding for
  face recognition and clustering}. In: IEEE Conference on Computer Vision and
  Pattern Recognition (2015)

\end{thebibliography}

\end{document}


\appendix
\section{Implementation Detail}
Figure~\ref{fig_overview} shows an overview of the whole procedure.

\begin{figure}[!ptb]
\centering
\includegraphics[height=\textheight]{figures/overview.png}
\caption{Overview of the whole procedure.}
\label{fig_overview}
\end{figure}

In Section~\ref{data_preparation}, 750 matches of data are split into 3,238 phases having total of 28,830 phases. Phases with a length of 10 minutes or less were absorbed into adjacent phases to ensure the minimum duration. Those with more than 8 measured players are used for role learning and player-wise role clustering in Section~\ref{data_labeling}. As a result, 17,953 player-phase entities measured from 1,989 merged phases are used in this study, each of which is labeled as one of 657 player-role entities generated from 436 players.

For training-validation-test split in Section~\ref{heatmap_gen}, 308 player-role entities with more than 20 phases in the dataset are used to construct the test set by sampling 10 phases for each entity. The validation set is constructed by sampling 5 phases for each of 332 player-role entities having more than 15 phases in the remaining data. That is, the dataset of size 17,953 (player-phase entities) are split into 13,213 for training, 1,660 for validation, and 3,080 for test.

The augmentation tasks in Section~\ref{heatmap_aug} are separately performed for the training, validation, and test datasets. They result in the augmented training data of size 49,519 (heatmap pairs), validation data of size 3,320, and test data of size 36,960. (Note that the validation and test data is augmented by the exhaustive 3-combination. Thus, the resulting dataset sizes are $\binom{5}{3} = 10$ and $\binom{10}{3} = 120$ times the number of corresponding player-role entities, respectively.)

\begin{table}[!htb]
\caption{The locational and directional branch CNNs of 6MapNet have the same structure as in this table. Batch normalization is done after every layer, and dropouts with probatility 0.25 are done after some layers.} \label{tab_cnn}
\centering
\vspace{1em}
{\renewcommand{\arraystretch}{1.2}
\renewcommand{\tabcolsep}{0.1cm}
\begin{tabular}{|c|c|c|c|c|c|}
\hline
Layer & Input & Padding & Kernel & Activation & Output\\
\hline
Conv1a & $35 \times 50 \times 1$ & $1 \times 0$ & $2 \times 3 \times 4$, 1 & ReLU & $36 \times 48 \times 4$\\
\hline
Conv1b & $36 \times 48 \times 4$ & $1 \times 1$ & $3 \times 3 \times 4$, 1 & ReLU & $36 \times 48 \times 4$\\
\hline
MaxPool1 & $36 \times 48 \times 4$ & - & $2 \times 2$, 2 & - & $18 \times 24 \times 4$\\
\hline
Conv2a & $18 \times 24 \times 4$ & $1 \times 1$ & $3 \times 3 \times 16$, 1 & ReLU & $18 \times 24 \times 16$\\
\hline
Conv2b & $18 \times 24 \times 16$ & $1 \times 1$ & $3 \times 3 \times 16$, 1 & ReLU & $18 \times 24 \times 16$\\
\hline
MaxPool2 & $18 \times 24 \times 16$ & - & $2 \times 2$, 2 & - & $9 \times 12 \times 16$\\
\hline
Conv3a & $9 \times 12 \times 16$ & $1 \times 1$ & $2 \times 3 \times 32$, 1 & ReLU & $10 \times 12 \times 32$\\
\hline
Conv3b & $10 \times 12 \times 32$ & $1 \times 1$ & $3 \times 3 \times 32$, 1 & ReLU & $10 \times 12 \times 32$\\
\hline
MaxPool3 & $10 \times 12 \times 32$ & - & $2 \times 2$, 2 & - & $5 \times 6 \times 32$\\
\hline
Conv4a & $5 \times 6 \times 32$ & $1 \times 1$ & $3 \times 3 \times 64$, 1 & ReLU & $5 \times 6 \times 64$\\
\hline
Conv4b & $5 \times 6 \times 64$ & $1 \times 1$ & $3 \times 3 \times 64$, 1 & ReLU & $5 \times 6 \times 64$\\
\hline
FC1 & 1920 & - & - & ReLU & 128\\
\hline
FC2 & 128 & - & - & - & 10\\
\hline
\end{tabular}}
\end{table}

In Section~\ref{6mapnet}, 6MapNet takes triplets of heatmap pairs and embeds each heatmap pair to a 20-dimensional vector. (Table \ref{tab_cnn} shows the structure of a single branch CNN.) As in FaceNet~\cite{Schroff2015}, the choice of triplets has a great influence on the performance. Thus, we decided to select a random positive and a hard negative for each heatmap pair as an anchor. Specifically, the triplet selection is performed as follows for every 10 or fewer epochs:
\begin{enumerate}
\item[(a)] Sample 5 heatmap pairs for each identity to form a candidate set $S$.
\item[(b)] Use every heatmap pair in the dataset as an anchor $\textbf{x}^a$ and combine it with the corresponding 5 positive candidates $\textbf{x}^p \in S^a = \{ \textbf{x}^p \in S | y^p = y^a \}$ (i.e., $\textbf{x}^p \in S$ that is of the same identity as $\textbf{x}^a$) to make 5 positive pairs $(\textbf{x}^a, \textbf{x}^p)$.
\item[(c)] For each positive pair $(\textbf{x}^a, \textbf{x}^p)$, pick one hard negative $\textbf{x}^n \in S-S^a$ that does not satisfy Eq.~\ref{triplet_ineq}. If there is no such negative, randomly choose one of the 10 negatives with the smallest distance from the anchor.
\end{enumerate}

In particular, the rate of positive pairs with no hard negative in (c), i.e.,
\begin{equation}
\left| \lbrace (\textbf{x}^a, \textbf{x}^p) \in P : \Arrowvert f(\mathbf{x}_i^a) - f(\mathbf{x}_i^p) \Arrowvert_2^2 + \alpha \leq \Arrowvert f(\mathbf{x}_i^a) - f(\mathbf{x}_i^n) \Arrowvert_2^2 \ \forall \textbf{x}^n \in S-S^a \rbrace \right| / |P| \label{val_accuracy}
\end{equation}
(where $P$ is the set of all positive pairs selected in (b)) is used as validation accuracy during the training. As a result, 244,310 triplets for the training data and 14,940 for the validation data are selected for every 10 or fewer epochs.

We start the training with randomly initialized weigths. Adam is used as the optimizer with initial learning rate 0.05, which is decayed during the training procedure. We fix mini-batches of size 1,000, so the weights are updated 245 times per epoch. The model is trained with a maximum of 10 epochs for a selected triplets. If there is no significant increase in validation loss, the triplet is newly selected. During each triplet selection task, the learning performance is evaluated with the validation accuracy given in Eq.~\ref{val_accuracy}. If there is no improvement in the validation accuracy, the whole learning procedure is finished.

The training goes very quickly. It starts with nearly 0\% accuracy and reaches its peak performance after dozens of epochs. The best performance is obtained when $\alpha$ in Eq.~\ref{triplet_loss} is set to 0.1. We used the model with 94.5\% validation accuracy (in terms of Eq.~\ref{val_accuracy}) in the experiments of the next sections.